\documentclass[11pt,a4paper]{article}
\usepackage[hyperref]{conll-2019}

\usepackage{times}
\usepackage{url}
\usepackage{latexsym}
\usepackage{graphicx}
\usepackage[11pt]{moresize}

\aclfinalcopy 

\title{Learning to Answer Subjective, Specific Product-Related \\ Queries using Customer Reviews by Adversarial Domain Adaptation}

\author{\bf Manirupa Das, 
        \bf Zhen Wang, 
        \bf Evan Jaffe,
        \bf Madhuja Chattopadhyay, \\
        \bf Eric Fosler-Lussier 
        \bf \& Rajiv Ramnath \\
  The Ohio State University \\
  {\tt \{das.65, wang.9215, jaffe.59, chattopadhyay.34, }\\ 
  {\tt  fosler-lussier.1, ramnath.6\}@osu.edu} \\
  }
  
\date{}

\begin{document}
\maketitle
\begin{abstract}
  Online customer reviews on large-scale e-commerce websites, represent a rich and varied source of opinion data, often providing subjective qualitative assessments of product usage that can help potential customers to discover features that meet their personal needs and preferences. Thus they have the potential to automatically answer specific queries about products, and to address the problems of \textit{answer starvation} and \textit{answer augmentation} on associated consumer Q \& A forums, by providing good answer alternatives. In this work, we explore several recently successful neural approaches to modeling sentence pairs, that could better learn the relationship between questions and ground truth answers, and thus help infer reviews that can best answer a question or augment a given answer. In particular, we hypothesize that our adversarial domain adaptation-based approach, due to its ability to additionally learn domain-invariant features from a large number of unlabeled, unpaired question-review samples, would perform better than our proposed baselines, at answering specific, subjective product-related queries with reviews. We validate this hypothesis 
  using a small gold standard dataset of question-review pairs evaluated by human experts, surpassing our chosen baselines. Moreover, our approach, using no labeled question-review sentence pair data for training, gives performance at par with another method utilizing labeled question-review samples for the same task.
\end{abstract}
\section{Introduction}

General question-answering (QA), in the context of opinion and qualitative assessments available to consumers via Q \& A forums on product-based e-commerce websites, is a challenging open problem. For example, consider a real-world question such as: \textit{``Is the Canon EOS Rebel T5i worth the extra \$200+ dollars to get as a starter camera, or should I just go with the cheaper T3i?"}. Many such questions cannot be answered directly using knowledge bases constructed from product descriptions alone \cite{mcauley2016addressing}, but clearly rely on personal experiences of others, for a satisfactory answer. Some questions, especially on newer items  may not be immediately  answered -- which can lead to ``answer starvation''. Many questions have short, unrelated, or incomplete answers -- these are candidates for ``answer augmentation'', via plausible answer alternatives.

Product-related question answering (PRQA) has emerged as a new research area, different from traditional QA and community question answering (CQA) tasks, owing to the immense popularity of e-commerce websites, where answers to potential customer questions often involve opinions and experiences from different users, found in the plentiful customer reviews that can help discover products or features for a more personalized experience \cite{yu2018aware, wan2016modeling}. Besides a binary ``good" or ``bad" assessment, product reviews tend to provide a wide range of : (i) personal experiences;  (ii) subjective qualitative assessments, (iii) unique use-cases or failure scenarios. Moreover, massive volume and range of opinions makes review systems difficult to navigate \cite{wan2016modeling}. This opinion data raises two interesting questions: 
(i) How can we help users navigate massive volumes of consumer opinions to address ``specific queries''? (ii) How can we build an end-to-end system that simultaneously leverages the labeled question-answer data from QA forums, with abundant unlabeled reviews that may carry valuable information that can answer a specific product-related question?  Thus, inspired by \citet{wan2016modeling} and \citet{mcauley2016addressing}, and a plethora of available architectures, we define the task of learning to answer product-related questions with reviews as:\\ \textit{``To be able to respond to specific, subjective, product-related queries automatically with reviews, to address problems such as answer starvation and answer augmentation by providing suitable answer alternatives, leveraging available signal from answer sentence data to enable learning of relevant review sentences that can address a question, with minimal supervision."}

Given recent advances in sentence pair modeling  and question answering \cite{sharp2016creating, rocktaschel2015reasoning, yin2015abcnn} via end-to-end neural approaches \cite{rajpurkar2016squad, nguyen2016ms, sukhbaatar2015end}, we attempt to adapt an architecture that is most suitable for our problem setting.
One key idea to note in the two types of user-generated sentences used in our task is that answer sentences and review sentences, though both capable of addressing user questions, are both created with very different intent and purpose. The former is specifically targeted toward some or all aspects of a question, and the latter providing some personal experiences and qualitative assessments regarding a product that may or may not satisfy some user question. Thus answer and review sentences come from two very different distributions. We hypothesize therefore, that being able to automatically learn the notion of a ``good'' or ``correct'' answer, i.e. learn features that constitute such an answer by leveraging the commonalities between correct answers and suitable review sentences, might be key to our solution. In this context, 
we propose an adversarial product review--based question answering approach for our task in a minimally supervised setting, inspired by neural domain adaptation due to \citet{ajakan2014domain} and \citet{ganin2016domain}. To our knowledge, we are the first to address product-related question answering by identifying answers from unlabeled review data with no supervision signal on the reviews and deriving only weak supervision from labeled question-answer data, in an adversarial neural domain adaptation setting.

\section{Background and Related Work}

\subsection{Addressing subjective product-related queries with reviews}

\textbf\textit{{``Mixture-of-Experts"}} frameworks combine several weak learners
by aggregating their outputs with weighted confidence scores. In their 2016 work, McAuley \& Yang show that such a model can be adapted to simultaneously identify relevant reviews and combine them to answer complex queries, by treating reviews as experts that either support or oppose a particular response. 
Bi-linear models \cite{chu2009personalized} can help to address the issue of questions and reviews being from different domains drawing from very different vocabularies, by learning complex mappings between words in one corpus and words in another (or more generally between arbitrary feature spaces), which can be regarded as a form of domain adaptation.
\cite{mcauley2016addressing} thus develop a ``mixture-of-experts''-based bilinear model, called MoQA, to  simultaneously learn which customer opinions are relevant to the query, as well as a prediction function that allows each review opinion to 'vote' on the response, in proportion to its relevance. These relevance and prediction functions are learned automatically from large corpora of training queries and reviews .

\citet{yu2018aware} develop an answer prediction framework which consists of two components, viz., an aspect analytics model and a predictive answer model. Given a product category, the aim of the aspect analytics model is to detect and capture latent aspects from a collection of review texts in an unsupervised manner. To this end they employ a 3-order Autoencoder to model aspects from review texts in the same product category and learn aspect-specific embeddings for reviews. This aspect analytics model generates aspect distributions and embeddings of reviews capturing hidden semantic features associated with certain aspects. The predictive answer model captures intricate relationships among question texts, review texts, and yes-no answers reporting answer prediction numbers  surpassing to \cite{mcauley2016addressing}. However, this work caters only to $yes/no$ questions in the dataset being considered, which is the same as ours, and does not involve learning matching features for open-ended questions, which may have subjective, opinion-based answers, which are the main focus in our work. 

\citet{chen2019answer} propose an answer identification framework from reviews which employs a multi-task attentive network, called QAR-net, leveraging both large-scale user generated question-answer data and manually labeled question-review data to achieve this goal. Multi-task learning can be an effective learning paradigm for boosting the performance of tasks with insufficient training instances by training jointly with related tasks having abundant training data, and they couple this paradigm with Attention to obtain a network that allows a ``question focus'' to attend to various ``answer patterns'' across answer and review sentences. However they utilize manually labeled question-review samples for training purposes; thus it is not a completely unsupervised approach. Our work, however, being similar in setting to \cite{chen2019answer}, differs from works like MoQA in that MoQA uses reviews as supporting data for answer prediction (i.e. uses review sentences as supporting “experts” for “Yes” or “No” binary questions; ranking answers before non-answers for open-ended questions), and does not try to actually identify potential answers from review sentences. 
Another recent work, AdaMRC \cite{wang2019adversarial}, for the related task of machine reading comprehension (MRC), where only unlabeled passages are available in the target domain, leverages domain adaptation to more effectively alleviate noise arising from a data augmentation step. Here, synthetic question-answer pairs are first generated for passages in the target domain before adversarially training a domain classifier on pseudo-generated question-answers and human annotated question-answer pairs, outperforming state-of-the-art MRC systems such as SAN \cite{liu2017stochastic} and BiDAF \cite{tuasonbidaf} on various datasets.


\subsection{Choices for Modeling Sentence Pairs}

Text pairs can exhibit various relations, including paraphrase, entailment, question-answer, translation and more.
Early systems designed to model these relations based on lexical overlap or word pairs often fail to generalize to unseen word pairs, and can have difficulty learning synonymy effectively due to sparse features.
Neural networks with dense text embeddings can more effectively learn synonymy and other relations, with attention helping to extend learning beyond word pairs to phrasal pairs \cite{bahdanau2014neural}.

\subsubsection{BiCNN and ABCNN-3}
Convolutional neural networks (CNNs) have been used effectively for a variety of natural language processing tasks \cite{kim2014convolutional}, following earlier successes in image recognition \cite{krizhevsky2012imagenet}. 
The hierarchic nature of language is a natural parallel to local structure in image data, where nearby words often form meaningful phrases in the way nearby pixels form meaningful sub-units of the complete image.

The ABCNN \cite{yin2015abcnn} set of models is known to be effective at modeling sentence pair relations for tasks including answer selection, paraphrase identification, and textual entailment.
In our work, the most effective reported versions of ABCNN without and with attention viz. BiCNN and ABCNN-3 are selected for use as a baseline for answer- or review-selection. 
The baseline BiCNN used in this work consists of two weight-sharing CNNs, each processing one of the two sentences, and a final logistic regression layer at the top that solves the sentence pair task by making a sentence pair binary labeling decision.  

While the non-attention-based BiCNN model is shown to have performance comparable to the full ABCNN models with fewer parameters, we choose ABCNN-3 as an additional baseline to evaluate against, as it combines the strengths of their other two attention-based models ABCNN-1 and ABCNN-2, allowing the attention mechanism to operate both on the convolution and on the pooling parts of a convolution-pooling block in this architecture. We adapt these models for our work from a third-party implementation\footnote{http://github.com/galsang/ABCNN}.

%
%

\subsubsection{Reasoning for Textual Entailment}

Attention-based neural networks have recently demonstrated success in a wide range of tasks ranging from
handwriting synthesis \cite{graves2013generating}, machine translation \cite{bahdanau2014neural}, to image captioning \cite{xu2015show}, and speech recognition \cite{chorowski2015attention} to sentence summarization \cite{rush2015neural}. The idea is to allow the model to attend over past output vectors, 
thereby mitigating the LSTM’s cell state bottleneck. More precisely, an LSTM with attention for recognizing textual entailment (RTE) does not need to capture the whole semantics of the premise in its cell state. Instead, it is sufficient to output vectors
while reading the premise and accumulating a representation in the cell state that informs the second LSTM which of the output vectors of the premise it needs to attend over, to determine the RTE class \cite{rocktaschel2015reasoning}. We believe that this type of sentence pair model lends itself well to a QA task setting such as ours, where a potential answer may bear some degree of an entailment relation with a question. Hence we choose the RTE model due to \cite{rocktaschel2015reasoning} as another baseline model to evaluate against.


\subsection{Neural Domain Adaptation}

Top-performing deep neural architectures are trained on
massive amounts of labeled data. In the absence
of labeled data for a certain task, however, domain adaptation (DA)
often provides an attractive option given
that labeled data of similar nature but from a different
domain (e.g. synthetic images) are available. As the training progresses, the approach promotes
the emergence of ``deep" features that are
(i) discriminative for the main learning task on the source domain and (ii) invariant with respect to the shift between the domains. This adaptation behavior could be achieved in almost any feed-forward model by augmenting it with few standard layers and a simple new gradient reversal layer, according to works on this architecture, due to \cite{ajakan2014domain, ganin2015unsupervised}. 
The resulting augmented architecture is thus trained using standard back-propagation. 
While this approach has not traditionally been used in the literature for modeling sentence pairs, it presented an interesting choice, given our particular task setting of candidate answer sentences from two different but related data distributions, with one having human-generated labels and the other unlabeled with respect to customer questions. Given that this model was a good fit for the domain variance aspect of our problem, we considered how this model could be  adapted to our particular sentence pair modeling scenario.

\section{Adversarial Product Review--based Question Answering}

Given our task of predicting relevant reviews that can answer a specific product-related question, or provide additional detail for it, we  hypothesize that our task is well-suited for and can benefit from domain adaptation, in which the data at training and test time come from similar but different distributions. In our case, these distributions correspond to ground truth answers and unlabeled reviews, and here, we want to employ a representation learning approach for effective transfer of information in reviews that is ``related'' or ``well-matched'', to ``good answers'' to specific questions. For such transfer to be achieved, predictions must be made based on features that cannot discriminate between the training (source) and test (target) domains. Thus our proposed model for this work is the one adapted 
from \cite{ganin2016domain} but with the input as \textbf{sentence-pair data}, i.e. question-answer and question-review pairs, and we expect this model to do better than our other chosen sentence pair baselines. In our experiments the \textbf{labeled answer sentences} to questions represent the \textbf{source domain data} and unpaired (unlabeled) reviews represent the \textbf{target domain data}. Our proposed approach to domain adaptation is thus to train on large amounts of labeled data from the source domain (labeled Q-A pairs) and large amount of unlabeled (previously unpaired) data from the target domain, i.e reviews, as no labeled target domain data is necessary \cite{ganin2015unsupervised}. Our experiences with this model is further described in the Experiments section.


\begin{figure*} [ht!]
\centering
\includegraphics[scale=0.6]{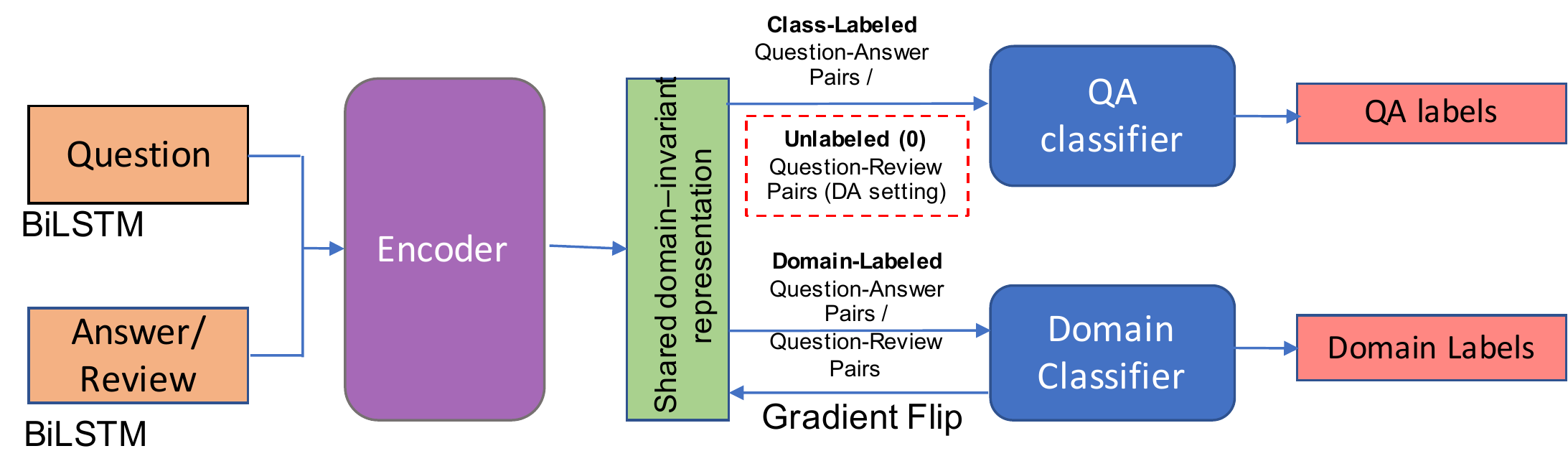}
\caption{Architecture of our model for Question-Answer/Review sentence pairs, adapted from the Domain Adversarial Neural Network due to \cite{ajakan2014domain, ganin2016domain} }
\label{tfdann}
\end{figure*}

\subsection{Domain Adversarial Neural Network}

The Domain Adversarial Neural Network (DANN) of \cite{ajakan2014domain, ganin2016domain}  performs domain adaptation by optimizing a minimax training objective that simultaneously learns to do well at task-specific label prediction while doing poorly at domain label prediction. This is motivated by the theory on domain adaptation \cite{ben2010theory} that a transferable feature is one for which an algorithm cannot learn to identify the domain of origin of the input observation. In the DANN model this is achieved by adversarial training of a domain classifier by reversing the gradient from it during backpropagation. Our adaptation of this model passes in \textit{labeled} question-answer (QA) pairs and \textit{unlabeled} question-review (QR) pairs thereby learning to answer a question and learning the domain-invariant features between two domains, i.e. review sentences and answer sentences, at the same time. To be specific, there is an encoder to learn the vector representation for the question and the answer/review separately. Afterwards, the representations are concatenated together to represent the paired data (``question + review" and ``question + answer"). Only ``question + answer" pairs are input into the label predictor which in our case is the \textbf{QA classifier}, to predict whether the answer could really answer the question. Both of ``question + answer" pairs and ``question + review" pairs are input into a \textbf{Domain classifier} to predict whether the current input is a answer or a review. Both, the QA classifier and Domain classifier use cross entropy loss for  back-propagation. The main idea of domain adaptation here is reversal of the gradient coming from the domain classifier as shown in Figure \ref{tfdann}. That is to say, when the gradients of domain classifier are back-propagated to the encoder, their negative values are actually used to update the parameters of the encoder. The point is to eliminate the influence of domain-specific features learned by the domain classifier and keep the domain-invariant features. Thus, in the inference stage, when a review is input into the QA classifier, the encoder will learn some features which are similar to answer domain and the QA classifier will predict whether the review can answer the question or not. We hypothesize that this approach can greatly help with answering open-ended product-related questions with reviews.

Our implementation of the DANN model uses separate bidirectional LSTMs to encode question and answer or review sentence separately to get their vector representation $V_{q}, V_{a/r}$ and concatenates them to get the vector representation for the pair, $[V_{q}; V_{a/r}]$. The inputs to the model are the 300 dimensional GloVe vectors with random uniform sampled vectors for unknown words. The hidden state for the paired LSTM is also 300-dimensional. The QA classifier and domain classifier are each two-layer (512-256) feed forward fully-connected networks. A gradient flip layer is added before the domain classifier. When calculating the loss, we use a mask operation to calculate the softmax loss of the QA classifier only for QA pair inputs.


\newpage

\section{Dataset}

We use the newer Q \& A dataset made publicly available by \cite{mcauley2016addressing}\footnote{http://jmcauley.ucsd.edu/data/amazon/}, the authors of the original work on answering product-related questions with reviews, which was developed off of the original SNAP dataset for Amazon product reviews and ratings \footnote{https://snap.stanford.edu/data/web-Amazon.html}. This dataset consists of paired questions and answers and de-duplicated product reviews on Amazon, across 24 different product categories. The question and answer data, total around \textbf{1.4 million answered questions}. 56.1\% of the questions are binary, i.e. having ``Yes'' or ``No'' answers, and the rest constitute ``Open-Ended'' questions with more subjective or specific answers. For our work we select only the set of \textbf{Open-Ended (OE)} questions that may include \textbf{multiple} answers to each question, from 6 product categories, viz. Automotive, Baby, Electronics, Home \& Kitchen, Sports \& Outdoors and Tools and Home Improvement. Product reviews and metadata from Amazon, total 142.8 million spanning May 1996 - July 2014, and includes reviews (e.g. ratings, text, helpfulness votes), and product metadata (e.g. descriptions, category). After cleaning and matching by product id (ASIN), we have a total of \textbf{128K unique ASINs}, that match a total of \textbf{1,06,1402 OE} Q \& A pairs, with \textbf{2,852,954 reviews}, as shown in Table \ref{dataset}. Important to note here is that the datasets of QA pairs are created in a way such that review answer sentences are only selected from the matching subset of reviews corresponding to the same product from the category that the question belongs to. We also have a small hand-labeled dataset of \textbf{1725 Question-Review pairs} acquired from the authors of the original work \cite{mcauley2016addressing} that corresponds to approximately 300 pairs for each of these 6 categories. This data was generated by human experts for evaluating their models, which we use for automated target domain-only evaluation of our models. 

\begin{table}
\begin{center}
\begin{tabular}{|l| r | r |}
\hline 
\bf Category & \bf Q-A pairs & \bf Reviews \\
\hline
Automotive & 18,214 & 20,474 \\
Baby & 40,429 & 160,793 \\
Electronics & 472,678 & 1,689,189 \\
Home \& Kitchen & 283,637 & 551,683 \\
Sports \& Outdoors & 140,120 & 296,338 \\
Tools \& Home & 106,324 & 134,477 \\
\hline
\bf Total & \bf 1,061,402 & \bf 2,852,954 \\
\hline
\end{tabular}
\end{center}
\caption{Dataset Statistics for \textbf{Open-Ended, Multi-Answer} Q-A Pairs and Reviews matched on 128K unique ASINs from the Amazon product review dataset \cite{mcauley2016addressing}.\label{dataset}}
\end{table}


\section{Experiments}

In order to evaluate the hypothesis that domain adaptation improves candidate review sentence answer selection, this work evaluates a variety of models for our defined task, with and without domain adaptation.
Two baselines for sentence pair modeling are evaluated. These include: (1) a CNN-based Siamese network architecture due to \cite{yin2015abcnn} -- we experiment with both, their non-attentional BiCNN and attention-based ABCNN-3 models, and (2) a Reasoning for Textual Entailment (RTE) model employing a conditional encoding-based attentive LSTM architecture \cite{rocktaschel2015reasoning}. 
Question-answer data and a small set of question-review data from \cite{mcauley2016addressing} are used to train and evaluate the models respectively.
%
Given gold answers and reviews for product-related queries, the \textbf{answers} are taken as the \textbf{Source domain} and \textbf{reviews} are taken as the \textbf{Target domain}. Since our datasets for training and evaluation are fairly balanced, we report the best \textbf{accuracy} for evaluations with each of our models. Table \ref{balance} shows the proportion of 0/1 labels in each of our datasets. Each system was trained on labeled question-answer pairs for \textbf{Auto}, \textbf{Baby} and \textbf{Electronics} product categories and a larger \textbf{Combined} dataset involving each of the 6 categories listed in Table \ref{dataset}, for a total of 4 trained models per system, and scored on 4 sets of evaluations, one each on source and target domain test sets for each model.
\begin{table*}[t]
\begin{center}
\begin{tabular}{| l | c | c | c |}
\hline 
\bf Category  & \bf Training Set & \bf Eval Source Data & \bf Eval Target Data\\
\hline
Auto  & 
49.93\%/50.07\%
 &
50.01\%/49.99\%
 & 
 (300 total); 66.67\%(0)/33.33\%3(1)\\
 \hline
Baby  & 
49.86\%/50.14\%
 & 
49.86\%/50.14\%
 & 
 (300 total); 66.67\%(0)/33.33\%(1)\\
 \hline
 Electronics  & 
49.84\%/50.16\% 
&
49.85\%/50.15\%
& 
(252 total); 66.67\%(0)/33.33\%(1)\\
\hline
Combined  & 
80.22\%/19.78\%
&
50\%/50\%
&  
(1725 total); 
77.22\%(0)/22.78\%(1) \\
\hline
\end{tabular}
\end{center}
\caption{Label Proportions (0/1 for unrelated/related) in each dataset for a total number of instances\label{balance}.}
\end{table*}

\begin{table*}[t]
\begin{center}
\begin{tabular}{| l | c | c | c | c | c |}
\hline 
\bf Model on  & \bf Train & \bf Eval & \bf Attn. LSTM & \bf BiCNN & \bf ABCNN-3 \\
\bf Data & \bf Data & \bf Data & \bf Accuracy & \bf Accuracy & \bf Accuracy \\
\hline

Auto (Eval on Source) & Source & Source & 50.12\% & \bf 70.0\% & \bf 70.0\%\\
Auto (Eval on Target) & Source & \bf Target & \bf 66.67\%* & 53.0\% & 52.9\%*\\
\hline
Baby (Eval on Source) & Source & Source & 50.37\% & 71.0\% & \bf 71.4\%\\
Baby (Eval on Target) & Source &  \bf Target &  \bf 61.33\% & 54.0\%* & 51.4\%\\
\hline
Electronics (Eval on Source) & Source & Source & 60.41\% & \bf 74.0\% & 72.08\%\\
Electronics (Eval on Target) & Source & \bf Target & \bf 66.67\%* & 53.0\% & 52.05\%\\
\hline
Combined (Eval on Source) & Source & Source & 52.34\% & 69.0\% & \bf 70.55\% \\
Combined (Eval on Target) & Source & \bf Target & 53.73\% & \bf 64.0\%* &  61.44\%*\\
\hline
\end{tabular}
\end{center}
\caption{Results from Experiments with the Baseline Sentence-Pair models:  Conditional-encoding-based Attentive LSTM, BiCNN and ABCNN-3. \textbf{Bold} indicate best performance in that row of results while \textbf{*} indicates best performing for a particular model on target domain evaluation across individual categories or combined.\label{BaselineExpts} 
}
\end{table*}

\section{Results and Discussion}

\subsection{ABCNN models: BiCNN and ABCNN-3}
ABCNN results provide a baseline evaluation for question answering that does not involve domain adaptation. Because labeled sentence pairs are required for training, there is no easy way to incorporate unlabeled review data, which is plentiful but not targeted to any specific question. Results for both the BiCNN and ABCNN-3 models in Table~\ref{BaselineExpts} show strong performance when evaluated on the in-domain answer data, but a marked decrease when evaluated on out-of-domain review data, with BiCNN still giving the best performance on target domain evaluation for the Combined dataset at \textbf{64.0\%}. 

\subsection{Attentive LSTM for RTE model}

For modeling sentence pairs using textual entailment, we adapt the version of the conditional encoding-based attentive LSTM neural architecture of \cite{rocktaschel2015reasoning}, that uses the encoding for the question representation learnt by the first LSTM, with the cell state of the first LSTM, as conditional input into the second LSTM that models the answer or review sentence, and which then learns an attended representation over the conditional input to generate the final representation for QA classification\footnote{Our implementation of this model is taken from a third-party implementation -- https://github.com/shyamupa/snli-entailment for the same, and adapted to work for our dataset, and QA-based class labels instead of entailment.} \footnote{In order to run our experiments with this model, we tweaked the format of our dataset to get it to resemble SNLI format, but in our case we only had binary 'entailment' and 'contradiction' labels to represent 'yes' and 'no' answers and had no 'neutral' class label in the data. Both question-answer and question-review pairs were prepared in this way for passage through this model.}. Table \ref{BaselineExpts} outlines the results from the experiments with this model by individual category and also with the Combined training dataset for the source-only and target-only task, using the conditional encoding-based LSTM with attention adapted from the reasoning for textual entailment task. Our experiments show that the RTE model gives the best performance on target-domain review data for each individual category, getting \textbf{66.67\%}, \textbf{61.33\%} and \textbf{66.67\%} on Auto, Baby and Electronics categories, but with ABCNN models doing much better on target domain evaluations for the Combined dataset. It is perhaps worth noting that the RTE model consistently fares \textit{worse} on the source domain task when trained on source domain data as compared to out-of-domain target classification task, across categories, with no domain adaptation. This indicates perhaps that the conditional attention-based mechanism of this RTE model is able to better generalize for the out-of-domain task in the absence of domain adaptation.

\begin{table*}[ht!]
\begin{center}
\begin{tabular}{| l | c | c | c |}
\hline 
\bf DANN Model on Data & \bf Train Data & \bf Eval Data & \bf Accuracy\\
\hline
Auto (Eval on Source) & Source (QA pairs) & Source & 64.55\% \\
Auto (Eval on Target) & Source (QA pairs) & Target & 37.33\% \\
\hline
Baby (Eval on Source) & Source (QA pairs) & Source & 72.19\% \\
Baby (Eval on Target) & Source (QA pairs) & Target & 35.66\% \\
\hline
Electronics (Eval on Source) & Source (QA pairs) & Source & 75.19\% \\
Electronics (Eval on Target) & Source (QA pairs) & Target & 40.47\% \\
\hline
Combined 
(No domain-adapt) & Source (QA pairs) & Source & 83.28\% \\
Combined 
(No domain-adapt) & Source (QA pairs) & \bf Target & \bf 50.11\% \\
\hline
Combined 
\bf (Domain-adapt) & Source, Target (QA, QR pairs) & Source & 73.95\%* \\
Combined 
\bf (Domain-adapt) & Source, Target (QA, QR pairs) & \bf Target & \bf 77.17\%* \\
\hline
\end{tabular}
\end{center}
\caption{Results from experiments with the DANN \label{DANNtab} model. \textbf{Bold *} indicate best results with domain adaptation on target domain evaluations}
\end{table*}

\begin{table}[hb!]
\begin{center}
\begin{tabular}{|p{3cm}|p{4cm}|}
\hline 
\bf Question & \bf Review Sentence\\
\hline
{\small Since the hooks attach with velcro, do they \textbf{slide} or do they \textbf{stay in place} ?} & 
{\small ``I originally purchased the Mommy Hooks for our stroller and loved the durability of the metal, but ended up hating how big and clunky they are, and they are \textbf{not stationary}, \textbf{always sliding around}."} \\
\hline
{\small Is this \textbf{fit} with \textbf{2002 Camry} 2.4 L ?} &
{\small ``This filter was a \textbf{drop in replacement} for the air filter in my \textbf{2002 Camry} LE, V6."}\\
\hline

{\small Does {\textbf{this device}} offer sure protection ? Can it be attached to a {\textbf{surge protector}} ? Any problems using a {\textbf{powerstrip}} with it ( for electronics) ?}
&
{\small ``I mean this should go directly to your outlet and you can plug the {\textbf{surge protector power strip}} into {\textbf{this item}} and all of your devices into the \textbf{{surge protector strip}}.''} \\
\hline

\end{tabular}
\end{center}
\caption{Examples of Q-R target-only inference on positive examples that the DANN model gets right. \label{dannexamples}}
\end{table}

\subsection{DANN model}

The results from experiments with running the DANN model are listed in Table \ref{DANNtab}. When we do not use domain adaptation, the DANN is simply a binary QA classification system to answer product-related queries, trained only on Q-A pairs. When using domain adaptation, the model is trained on both Q-A and Q-R pairs, indicated by the red dotted box in Figure \ref{tfdann}. We evaluate our final model on source and target test sets, for Q-A pairs and Q-R pairs. As we can see from the Table \ref{DANNtab}, after using domain adaptation, the performance on Q-R pairs is improved greatly which demonstrates that adding Q-R pairs via domain adversarial training can help answer product queries using out-of-domain review sentences. 
As seen in the experimental results from our baseline sentence pair models and our proposed version of the DANN model in Tables \ref{BaselineExpts} and \ref{DANNtab}, we see that the DANN model outperforms the baselines on the out-of-domain Q-R pair classification task, validating our original hypothesis. The best performing ABCNN model evaluated on the source domain classification task gets a \textbf{70.55\%} accuracy on a held out test set of 10K Q-A samples, and \textbf{64.0\%} on Q-R classification with our human-labeled evaluation-only dataset from McAuley et al., of 1740 samples. Somewhat surprising is that our best performing RTE-based model gave a best accuracy of \textbf{66.67\%} 
for Electronics and Auto categories on out-of-domain Q-R classification, with Q-R classification significantly surpassing Q-A classification performance on all categories including Combined when trained on source domain data for RTE. Our DANN-based model was able to beat both of these baselines with an accuracy of \textbf{77.17\%} on Q-R classification, with large amount of \textbf{unlabeled} reviews incorporated into the domain adversarial training. Table \ref{dannexamples} shows some examples from Q-R inference using the DANN model, where it learned to correctly classify relevant review sentences that can answer a question. 
In addition, as seen in Table \ref{syscompare}, when compared with the \textbf{QAR-Net} system \cite{chen2019answer} for the same task, our method gives comparable performance on F-1 score and significantly higher precision, but for a combined dataset (1 million+ samples with all the categories) and trained with \textbf{unlabeled} question-review samples using domain adaptation, compared to data from only Electronics and Cellphones \& Accessories categories for QAR-Net, which uses labeled question-review instances for training. We believe this demonstrates the suitability and promise of our approach compared to other methods for this task.

\begin{table}[ht!]
\begin{center}
\begin{tabular}{| l | c | c | c |}
\hline 
\bf System & \bf Precision & \bf Recall & \bf F-1 score\\
\hline
QAR-Net & 53.85\% & \bf 60.67\% & 57.05\%\\
DANN & \bf 64.11\% & 52.46\% & 56.23\% \\
\hline
\end{tabular}
\end{center}
\caption{Comparison of QAR-Net system with the DANN model \label{syscompare} for PRQA. }
\end{table}

\section{Conclusion and Future work}
Our proposed adversarial domain adaptation approach for  question-answer/review sentence pair classification via domain adversarial training shows good results for learning to answer specific customer questions with product reviews. It is able to leverage plentiful unlabeled review data during training to better generalize to review data at inference time, significantly outperforming numerous baseline models that cannot easily incorporate such  data. Future work may involve incorporating unsupervised objectives on review data to further improve the model. Our novel approach to the PRQA task thus leverages previously unused data without requiring explicit supervision, using domain adaptation 
to identify specific product-related answers from product reviews. 

\bibliographystyle{apa}
\bibliography{conll-2019}

\begin{thebibliography}{}

\bibitem[\protect\astroncite{Ajakan et~al.}{2014}]{ajakan2014domain}
Ajakan, H., Germain, P., Larochelle, H., Laviolette, F., and Marchand, M.
  (2014).
\newblock Domain-adversarial neural networks.
\newblock {\em arXiv preprint arXiv:1412.4446}.

\bibitem[\protect\astroncite{Bahdanau et~al.}{2014}]{bahdanau2014neural}
Bahdanau, D., Cho, K., and Bengio, Y. (2014).
\newblock Neural machine translation by jointly learning to align and
  translate.
\newblock {\em arXiv preprint arXiv:1409.0473}.

\bibitem[\protect\astroncite{Ben-David et~al.}{2010}]{ben2010theory}
Ben-David, S., Blitzer, J., Crammer, K., Kulesza, A., Pereira, F., and Vaughan,
  J.~W. (2010).
\newblock A theory of learning from different domains.
\newblock {\em Machine learning}, 79(1-2):151--175.

\bibitem[\protect\astroncite{Chen et~al.}{2019}]{chen2019answer}
Chen, L., Guan, Z., Zhao, W., Zhao, W., Wang, X., Zhao, Z., and Sun, H. (2019).
\newblock Answer identification from product reviews for user questions by
  multi-task attentive networks.
\newblock {\em Proceedings of AAAI}.

\bibitem[\protect\astroncite{Chorowski et~al.}{2015}]{chorowski2015attention}
Chorowski, J.~K., Bahdanau, D., Serdyuk, D., Cho, K., and Bengio, Y. (2015).
\newblock Attention-based models for speech recognition.
\newblock In {\em Advances in Neural Information Processing Systems}, pages
  577--585.

\bibitem[\protect\astroncite{Chu and Park}{2009}]{chu2009personalized}
Chu, W. and Park, S.-T. (2009).
\newblock Personalized recommendation on dynamic content using predictive
  bilinear models.
\newblock In {\em Proceedings of the 18th international conference on World
  wide web}, pages 691--700. ACM.

\bibitem[\protect\astroncite{Ganin and Lempitsky}{2015}]{ganin2015unsupervised}
Ganin, Y. and Lempitsky, V. (2015).
\newblock Unsupervised domain adaptation by backpropagation.
\newblock In {\em International Conference on Machine Learning}, pages
  1180--1189.

\bibitem[\protect\astroncite{Ganin et~al.}{2016}]{ganin2016domain}
Ganin, Y., Ustinova, E., Ajakan, H., Germain, P., Larochelle, H., Laviolette,
  F., Marchand, M., and Lempitsky, V. (2016).
\newblock Domain-adversarial training of neural networks.
\newblock {\em Journal of Machine Learning Research}, 17(59):1--35.

\bibitem[\protect\astroncite{Graves}{2013}]{graves2013generating}
Graves, A. (2013).
\newblock Generating sequences with recurrent neural networks.
\newblock {\em arXiv preprint arXiv:1308.0850}.

\bibitem[\protect\astroncite{Kim}{2014}]{kim2014convolutional}
Kim, Y. (2014).
\newblock Convolutional neural networks for sentence classification.
\newblock {\em arXiv preprint arXiv:1408.5882}.

\bibitem[\protect\astroncite{Krizhevsky et~al.}{2012}]{krizhevsky2012imagenet}
Krizhevsky, A., Sutskever, I., and Hinton, G.~E. (2012).
\newblock Imagenet classification with deep convolutional neural networks.
\newblock In {\em Advances in neural information processing systems}, pages
  1097--1105.

\bibitem[\protect\astroncite{Liu et~al.}{2017}]{liu2017stochastic}
Liu, X., Shen, Y., Duh, K., and Gao, J. (2017).
\newblock Stochastic answer networks for machine reading comprehension.
\newblock {\em arXiv preprint arXiv:1712.03556}.

\bibitem[\protect\astroncite{McAuley and Yang}{2016}]{mcauley2016addressing}
McAuley, J. and Yang, A. (2016).
\newblock Addressing complex and subjective product-related queries with
  customer reviews.
\newblock In {\em Proceedings of the 25th International Conference on World
  Wide Web}, pages 625--635. International World Wide Web Conferences Steering
  Committee.

\bibitem[\protect\astroncite{Nguyen et~al.}{2016}]{nguyen2016ms}
Nguyen, T., Rosenberg, M., Song, X., Gao, J., Tiwary, S., Majumder, R., and
  Deng, L. (2016).
\newblock Ms marco: A human generated machine reading comprehension dataset.
\newblock {\em arXiv preprint arXiv:1611.09268}.

\bibitem[\protect\astroncite{Rajpurkar et~al.}{2016}]{rajpurkar2016squad}
Rajpurkar, P., Zhang, J., Lopyrev, K., and Liang, P. (2016).
\newblock Squad: 100,000+ questions for machine comprehension of text.
\newblock {\em arXiv preprint arXiv:1606.05250}.

\bibitem[\protect\astroncite{Rockt{\"a}schel
  et~al.}{2015}]{rocktaschel2015reasoning}
Rockt{\"a}schel, T., Grefenstette, E., Hermann, K.~M., Ko{\v{c}}isk{\`y}, T.,
  and Blunsom, P. (2015).
\newblock Reasoning about entailment with neural attention.
\newblock {\em arXiv preprint arXiv:1509.06664}.

\bibitem[\protect\astroncite{Rush et~al.}{2015}]{rush2015neural}
Rush, A.~M., Chopra, S., and Weston, J. (2015).
\newblock A neural attention model for abstractive sentence summarization.
\newblock {\em arXiv preprint arXiv:1509.00685}.

\bibitem[\protect\astroncite{Sharp et~al.}{2016}]{sharp2016creating}
Sharp, R., Surdeanu, M., Jansen, P., Clark, P., and Hammond, M. (2016).
\newblock Creating causal embeddings for question answering with minimal
  supervision.
\newblock {\em arXiv preprint arXiv:1609.08097}.

\bibitem[\protect\astroncite{Sukhbaatar et~al.}{2015}]{sukhbaatar2015end}
Sukhbaatar, S., Weston, J., Fergus, R., et~al. (2015).
\newblock End-to-end memory networks.
\newblock In {\em Advances in neural information processing systems}, pages
  2440--2448.

\bibitem[\protect\astroncite{Tuason et~al.}{}]{tuasonbidaf}
Tuason, R., Grazian, D., and Kondo, G.
\newblock Bidaf model for question answering.
\newblock {\em Table III EVALUATION ON MRC MODELS (TEST SET). Search Zhidao
  All}.

\bibitem[\protect\astroncite{Wan and McAuley}{2016}]{wan2016modeling}
Wan, M. and McAuley, J. (2016).
\newblock Modeling ambiguity, subjectivity, and diverging viewpoints in opinion
  question answering systems.
\newblock In {\em Data Mining (ICDM), 2016 IEEE 16th International Conference
  on}, pages 489--498. IEEE.

\bibitem[\protect\astroncite{Wang et~al.}{2019}]{wang2019adversarial}
Wang, H., Gan, Z., Liu, X., Liu, J., Gao, J., and Wang, H. (2019).
\newblock Adversarial domain adaptation for machine reading comprehension.
\newblock {\em arXiv preprint arXiv:1908.09209}.

\bibitem[\protect\astroncite{Xu et~al.}{2015}]{xu2015show}
Xu, K., Ba, J., Kiros, R., Cho, K., Courville, A., Salakhudinov, R., Zemel, R.,
  and Bengio, Y. (2015).
\newblock Show, attend and tell: Neural image caption generation with visual
  attention.
\newblock In {\em International Conference on Machine Learning}, pages
  2048--2057.

\bibitem[\protect\astroncite{Yin et~al.}{2015}]{yin2015abcnn}
Yin, W., Sch{\"u}tze, H., Xiang, B., and Zhou, B. (2015).
\newblock Abcnn: Attention-based convolutional neural network for modeling
  sentence pairs.
\newblock {\em arXiv preprint arXiv:1512.05193}.

\bibitem[\protect\astroncite{Yu and Lam}{2018}]{yu2018aware}
Yu, Q. and Lam, W. (2018).
\newblock Review-aware answer prediction for product-related questions
  incorporating aspects.
\newblock In {\em Proceedings of the Eleventh ACM International Conference on
  Web Search and Data Mining}, pages 691--699. ACM.

\end{thebibliography}
\end{document}